\documentclass[10pt,twocolumn,letterpaper]{article}

\usepackage{cvpr}              %

\usepackage[dvipsnames]{xcolor}

\definecolor{cvprblue}{rgb}{0.21,0.49,0.74}
\usepackage[pagebackref,breaklinks,colorlinks,citecolor=cvprblue]{hyperref}
\usepackage{catchfile} %
\usepackage{adjustbox}
\usepackage{multirow}
\usepackage{colortbl}

\usepackage[accsupp]{axessibility} %

\usepackage{lipsum}  
\usepackage{array}
\usepackage{makecell}
\usepackage{soul}
\newcolumntype{H}{>{\setbox0=\hbox\bgroup}c<{\egroup}@{}}
\usepackage{color}
\definecolor{light}{rgb}{0.5, 0.5, 0.5}
\def\light#1{{\color{light}#1}}

\def\paperID{10} %
\def\confName{CVPR}
\def\confYear{2024}

\title{Exploring the Benefits of Vision Foundation Models \\ for Unsupervised Domain Adaptation}

\author{Brunó B. Englert$^{*}$ \quad Fabrizio J. Piva$^{*}$ \quad Tommie Kerssies \quad Daan de Geus  \quad Gijs Dubbelman \\
Eindhoven University of Technology \\
{\tt\small \{b.b.englert, f.j.piva, t.kerssies, d.c.d.geus, g.dubbelman\}@tue.nl}
}

\begin{document}
\maketitle

\def\thefootnote{*}\footnotetext{Both authors contributed equally.}\def\thefootnote{\arabic{footnote}}

\begin{abstract}

Achieving robust generalization across diverse data domains remains a significant challenge in computer vision. This challenge is important in safety-critical applications, where deep-neural-network-based systems must perform reliably under various environmental conditions not seen during training. Our study investigates whether the generalization capabilities of Vision Foundation Models (VFMs) and Unsupervised Domain Adaptation (UDA) methods for the semantic segmentation task are complementary. Results show that combining VFMs with UDA has two main benefits: (a) it allows for better UDA performance while maintaining the out-of-distribution performance of VFMs, and (b) it makes certain time-consuming UDA components redundant, thus enabling significant inference speedups. Specifically, with equivalent model sizes, the resulting VFM-UDA method achieves an 8.4$\times$ speed increase over the prior non-VFM state of the art, while also improving performance by +1.2 mIoU in the UDA setting and by +6.1 mIoU in terms of out-of-distribution generalization. Moreover, when we use a VFM with 3.6$\times$ more parameters, the VFM-UDA approach maintains a 3.3$\times$ speed up, while improving the UDA performance by +3.1 mIoU and the out-of-distribution performance by +10.3 mIoU. These results underscore the significant benefits of combining VFMs with UDA, setting new standards and baselines for Unsupervised Domain Adaptation in semantic segmentation. The implementation is available at \href{https://github.com/tue-mps/vfm-uda}{https://github.com/tue-mps/vfm-uda}.

\end{abstract}
    
\section{Introduction}
\label{sec:intro}

\begin{figure}
    \centering
    \includegraphics[width=1.0\columnwidth]{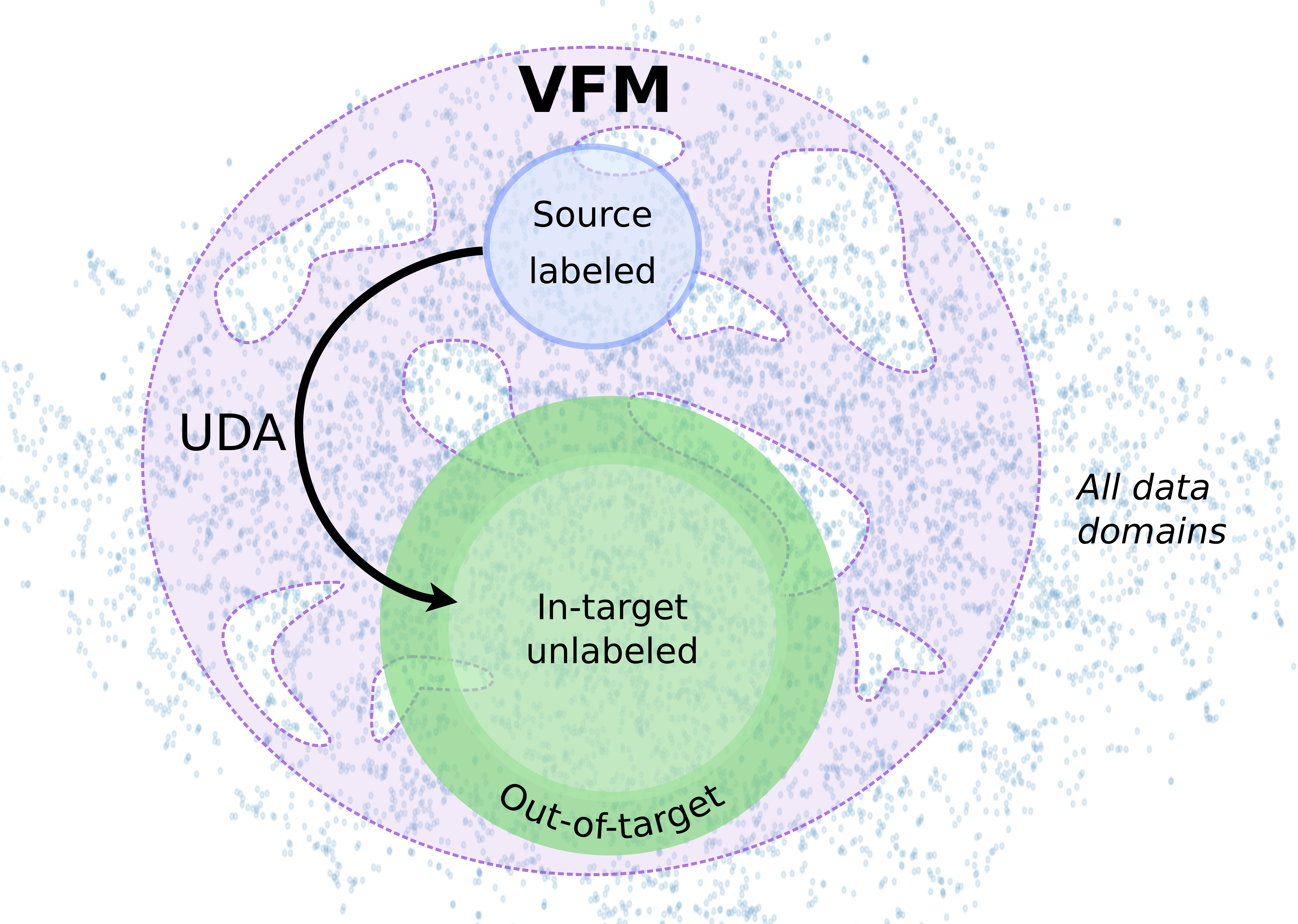}
    \caption{\label{fig:eye-catcher} \textbf{Generalization capabilities of UDA methods and VFMs.} UDA is designed to adapt a model from a labeled source domain to an unlabeled target domain, whereas VFMs capture a broad spectrum of data distributions, contributing to the overall generalization. The goal of this research is to investigate and leverage the combined in- and out-of-target generalization capabilities of UDA and VFMs.}
\end{figure}

In machine learning, generalization refers to a model's ability to perform well on data inside, near, and outside the distribution of the data on which it was trained. The challenge of achieving good generalization increases with the distance to the training data distribution. To maximize the chance of good generalization in real-world environments, models are best trained on a diverse dataset having a broad distribution, thus minimizing the likelihood of encountering out-of-distribution data. However, for dense tasks like semantic segmentation, obtaining abundant labeled data can be costly and labor intensive~\cite{cordts_cityscapes_2016}, as every pixel has to be labeled. As a result, annotated data is likely to be scarce, and networks trained on limited labeled data suffer from poor generalization due to the lack of exposure to sufficiently diverse training examples. To address the lack of generalization, Vision Foundation Models (VFMs)~\cite{dinov2_2023, eva02_2023, kirillov_sam_2023, mae_2022} and Unsupervised Domain Adaptation (UDA) ~\cite{mic_2023, hoyer_hrda_2022, xie_sepico_2022, cluda_2022} have emerged, amongst other alternatives~\cite{Peng_dgss_2022, Liu_udarev_2022,Yang_semisup_2023, Yuan_vfmpow_2023}. UDA methods leverage unlabeled data to adapt a model to a specific target domain, which is typically the domain where the model is to be deployed. By doing so, they do not necessarily aim for wide generalization beyond this target domain. In contrast, VFMs leverage extensive pre-training on large datasets to create models that can be used for efficient fine-tuning on various downstream tasks. Once fine-tuned on (limited) labeled data, these VFMs can generalize better~\cite{dinov2_2023,wei2024stronger} than models that were not as extensively pre-trained. In other words, UDA methods focus on performing well on a specific target domain, whereas VFMs can improve the generalization on domains that are unseen during fine-tuning. Both types of generalization are important, and they are illustrated in Fig.~\ref{fig:eye-catcher}. In this work, we study whether the generalization capabilities of UDA and VFMs are complementary.

Recently, Vision Foundation Models (VFMs) have made significant contributions by offering pre-trained models that excel in generalization, requiring minimal fine-tuning on downstream tasks~\cite{dinov2_2023, eva02_2023, radford_clip_2021, mae_2022}. Contrary to the traditional approach of pre-training models on labeled datasets like ImageNet~\cite{Russakovsky_imagenet_2015} or MSCOCO~\cite{Lin_mscoco_14}, Vision Foundation Models (VFMs) stand out by utilizing extensive pre-training on labeled and/or unlabeled datasets. Training on unlabeled data is done using different self-supervised techniques such as masked image modeling~\cite{Xie_simmim_22} or self-training~\cite{Caron_sslfm_2021}. Specifically in the context of semantic segmentation, VFMs have shown promising results in improving the performance to domains never seen during fine-tuning~\cite{wei2024stronger}.

Alternatively, Unsupervised Domain Adaptation (UDA) methods continue to make progress in enabling models to adapt to any unlabeled target domain~\cite{mic_2023, hoyer_hrda_2022, cluda_2022, xie_sepico_2022}. To adapt a model to a target domain, UDA methods use a labeled source domain, consisting of either synthetic images~\cite{richter_playing_nodate, ros_synthia_2016} or real images. The goal is to bridge the gap between the source and target domains, transferring what the model has learned from the source to perform as well as possible in the target domain. UDA methods are typically only evaluated on this target domain, but this does not reflect their generalization performance. Therefore, Piva \etal~\cite{Piva_uda_23} proposed to evaluate these models on an additional, unseen dataset, and they show that UDA methods can also improve the performance in this out-of-target setting.

Despite significant individual advancements by Vision Foundation Models (VFMs) and Unsupervised Domain Adaptation (UDA) methods, both have been studied in isolation, and it remains an open question to what extent they are complementary to each other. To address this gap in research, this work investigates the integration of VFM in UDA to obtain increased in- and out-of-target performance. For this purpose, we incorporate VFMs into a representative state-of-the-art UDA method, MIC~\cite{mic_2023}. We conduct ablations over components, image resolution, and self-training strategies, and assess the impact of VFM size and pre-training strategy. Based on these results, we adopt the best combination of VFM and UDA components which we refer to as the VFM-UDA method. The experimental results across synthetic-to-real and real-to-real scenarios demonstrate that VFMs can have a very positive influence on a UDA method's ability to perform well on both in- and out-of-target domains. This highlights the potential of their future use in combination with UDA methods.

In summary, the contributions of this work are:
\begin{itemize}
     \item A careful investigation of the complementarity of VFMs and UDA, resulting in new UDA standards and baselines for the VFM era.
     \item A detailed ablation over the necessity of UDA components when used in combination with VFMs, and an investigation of the influence of VFM model size and pre-training strategy.
     \item A broad experimental validation of VFM-UDA and comparison to non-VFM UDA methods on synthetic-to-real and real-to-real dataset combinations.
\end{itemize}

\section{Related Work}
\label{sec:relwork}

\textbf{Vision Foundation Models (VFMs)} have brought notable advancements in generalization within computer vision, being trained on large-scale data and adaptable for multiple downstream tasks. For instance, CLIP~\cite{radford_clip_2021} learns high-quality visual representations through contrastive learning~\cite{Chen_contlearn_2020} with large-scale image-text pairs. MAE~\cite{mae_2022} utilizes a masked image modeling framework for image pixel reconstruction. SAM~\cite{kirillov_sam_2023}, trained on a large-scale segmentation dataset, extracts features from images and prompts to predict single or multiple segmentation masks. EVA02~\cite{eva02_2023} applies masked image modeling to a CLIP model's visual features, offering a unique approach to visual representation learning. DINOv2~\cite{dinov2_2023}, on the other hand, is pre-trained on carefully curated datasets without explicit supervision, showcasing its self-supervised learning strength. Most VFMs currently rely on the plain Vision Transformer (ViT)~\cite{Dosovitskiy_vit_2021} architecture, which outputs single-scale features, posing a design challenge when integrating them with UDA, as is explained in the next section.

\textbf{Unsupervised Domain Adaptation (UDA)} methods aim to increase the performance of a model on a known target domain. This domain usually represents the environment where the model is likely to be deployed in the real world. These models can leverage unlabeled target data and labeled data from a source domain to increase a model's performance on a target domain. UDA methods leverage techniques like feature alignment~\cite{hoyer_daformer_2022, Luo_clan_19, Yan_mmd_17}, self-supervised learning~\cite{mic_2023, hoyer_hrda_2022, cluda_2022, hoyer_daformer_2022, Piva_uda_23} and data augmentation~\cite{li_bidirectional_2019, Piva_uda_23} to minimize the discrepancy between source and target domain distributions. Current UDA methods typically use hierarchical encoders, consisting of either Convolutional~\cite{He_resnet_2016, Simonyan_vgg_2015} or Transformer~\cite{xie_segformer_2021, Liu_swin_2021} blocks that yield multi-scale features to obtain optimal performance on small-scale objects, whereas VFMs produce single-scale features. As such, state-of-the-art UDA methods are not directly compatible with VFMs in an optimal manner. This work focuses on bridging this incompatibility and applying these UDA techniques to VFMs, assessing them outside the standard practice of initializing on ImageNet~\cite{Russakovsky_imagenet_2015}, and evaluating their effectiveness in adaptation settings that consider both in-target as well as out-of-target performance. 

To the best of our knowledge, leveraging the combined generalization capabilities of VFMs and UDA has not been explored, and filling this gap is what we aim for in our work.

\section{Methodology}
\label{sec:method}

This section outlines the adaptations made to align UDA strategies with Vision Transformer (ViT) architectures to enable the integration of UDA with VFMs, and introduces the motivation and design of our experiments.

\subsection{VFM-UDA}
\label{sec:method_udavfm}

\paragraph{UDA baseline.} As a baseline, we start from MIC~\cite{mic_2023}, a state-of-the-art Unsupervised Domain Adaptation (UDA) method. MIC utilizes a student-teacher framework~\cite{laine2017temporal, araslanov_self-supervised_2021, hoyer_hrda_2022, mic_2023, cluda_2022, xie_sepico_2022} with some additional UDA components. In the student-teacher framework, the teacher network generates pseudo labels for the target domain which are then used to supervise the student network. The student network uses a vanilla cross-entropy loss on the labeled source domain and on the unlabeled target domain using the pseudo labels generated by the teacher network. The teacher network is updated with an exponential moving average (EMA) using the student model's parameters. Below, we specify the additional components of this UDA method. In \cref{sec:results:components}, we assess the effectiveness of each of these components in combination with a VFM. For the final VFM-UDA model, we keep only the components that remain effective.

Feature Distance (FD)~\cite{hoyer_daformer_2022} is a UDA strategy that adds a Mean Squared Error (MSE) loss between the student's encoder output and those of a frozen pre-trained encoder. Minimizing this MSE loss encourages the student model to retain the features learned during pre-training, balancing the adaptation to new domain-specific features with the preservation of essential general features.

Masked Image Consistency (MIC)~\cite{mic_2023} is a UDA strategy that introduces an asymmetry between the teacher and student models by masking out parts of the original images for the student in the target dataset. This is achieved by randomly generating a patch mask and masking out different parts of each image. This masking forces the student model to infer from contextual information from the unmasked regions.

HRDA~\cite{hoyer_hrda_2022} is a model architectural change that is aimed at making high-resolution segmentation predictions in both UDA and conventional supervised learning~\cite{Hoyer_DG_2024}. This is achieved by conducting semantic segmentation on both high-resolution (HR) and low-resolution (LR) crops of an image. The resulting segmentation predictions for the LR and HR crops are fused by a learned scale-attention head. This fusion approach leverages the detailed information from HR crops and the broader context from LR crops, with the added drawback of having to do multiple forward passes for one image.

\paragraph{VFM encoder.}  We choose the DINOv2~\cite{dinov2_2023} VFM as the primary encoder on top of which we conduct UDA, but we also evaluate alternative VFMs in our experiments in \cref{sec:results:vfms}. The UDA baseline method uses the MiT-B5 encoder, which produces multi-scale features. However, because all performant VFMs use single-scale VIT-based encoders, we need to make an adaption to the MIC baseline. This adaption is performed in the decoder, as described in the next section. To ensure a fair comparison of the VFM-UDA method to the baseline, we use encoders with a roughly equal number of learnable parameters. Specifically, we use the ViT-B/14 encoder with 86M parameters, while MiT-B5 has 81M parameters.

\begin{figure}
    \centering
    \includegraphics[width=0.45\columnwidth]{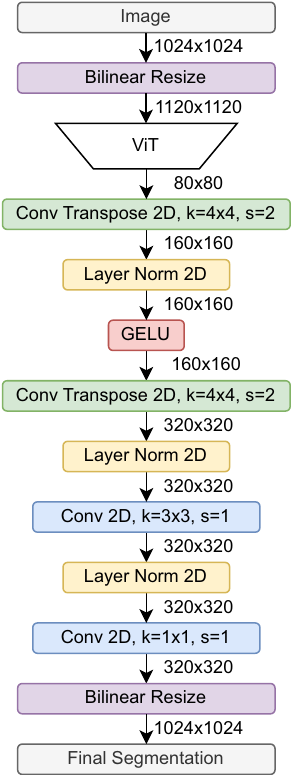}
    \caption{\textbf{Decoder head architecture.}\label{fig:decoder}}
\end{figure}

\paragraph{VFM decoder.} MIC and other well-performing UDA methods use decoders that are designed for encoders that output multi-scale features. However, ViTs only output single-resolution features.  This difference motivates us to use a different, yet much simpler decoder architecture that is specifically designed for ViTs. Our decoder, depicted in Fig.~\ref{fig:decoder}, is inspired by the Segment Anything Model's (SAM) \cite{kirillov_sam_2023} upsampling stage but is slightly modified for our use case. Compared to the SAM model's upsampling stage, we introduce an additional 3$\times$3 Conv2D before the final output. While a more complex and larger decoder head could be used, DINOv2 already performs well with a simple linear decoder and a frozen encoder \cite{dinov2_2023}. This suggests that a large decoder is not necessary for VFMs due to their extensive pre-training. The more efficient decoder for the VFM-UDA approach contains 1.8M parameters, in contrast to the MIC model's decoder, which has 5.2M parameters.

\paragraph{VFM masking.}  The baseline UDA method, MIC, uses direct image masking for the image masking consistency loss. In our approach, instead of applying a mask directly to the image, we mask the patch tokens and replace them with a learnable token, similarly to how BEiT is trained~\cite{bao_beit_2021}. This adjustment acknowledges the architectural differences in ViT models and optimizes the process for token-based architectures. 

\subsection{Experimental set-up}

We conduct several experiments to thoroughly assess the combined VFM-UDA method and support our design choices. 

\paragraph{Domain adaptation setup.} To assess the UDA capabilities of models, we evaluate the performance in synthetic-to-real and real-to-real adaptation scenarios. The synthetic-to-real scenario allows us to assess how well the model can bridge the gap between computer-generated images and real-world images, representing an extreme case of domain shift. On the other hand, real-to-real experiments test the model's ability to adapt between different real-world conditions, reflecting more subtle variations and complexities found in natural settings. This dual approach ensures a thorough evaluation of the VFM-UDA integration, highlighting its adaptability and performance across different visual domains.

\paragraph{In- and out-of-target evaluation.}
To truly assess a model's generalization capabilities, it should also be evaluated beyond the domain of its target dataset. Therefore, similarly to Piva \etal~\cite{Piva_genstdy_23}, we additionally evaluate each model on another dataset that falls outside of the distribution of the target domain. In other words, to measure a UDA method's performance both in-target and out-of-target, we use two completely separate evaluation datasets. This additional out-of-target dataset is never used during UDA training, to ensure there is no data leakage. Training and validation splits are made for both of these evaluation datasets. The training split for the out-of-target dataset is only used to determine the oracle performance.

\paragraph{Baseline, UDA, and oracle.} Our benchmark compares the UDA performance with respect to a \textit{baseline} and an \textit{oracle}. The baseline is trained on only the source domain, in a supervised manner, representing the model's performance before domain adaptation. The oracle is trained on only the labeled target data, also in a supervised manner, and reflects the empirical upper bound of a model's performance on the target domain. Both the baseline and the oracle rely only on supervised learning, meaning they do not use unlabeled data. In contrast, in the experiments with the UDA methods, we train on the source domain with labeled data and try to adapt to the target domain with unlabeled data, using different pre-training configurations and model sizes.

\paragraph{Datasets.} To assess the in-target performance, following previous state-of-the-art methods, we use GTA5~\cite{richter_playing_nodate} $\rightarrow$ Cityscapes~\cite{cordts_cityscapes_2016} as the synthetic-to-real scenario. For the real-to-real adaptation scenarios, we use Cityscapes $\rightarrow$ Mapillary~\cite{Neuhold_mapillary_17}. To assess the out-of-target performance, we use  WildDash2~\cite{Zendel_wilddash_18}, a completely separate dataset from the source and target datasets. We choose WildDash2 because it includes city, highway, and rural scenes under various weather conditions, and because the images are captured in more than 100 countries, providing diverse and challenging imagery.

\paragraph{Implementation details.} The encoder is a vanilla ViT-B/14~\cite{Dosovitskiy_vit_2021}  with DINOv2~\cite{dinov2_2023} pre-training. The learning rate for the decoder is $1.4\times10^{-4}$ and for the encoder it is $1.4\times10^{-5}$. We train for 40,000 iterations with a batch size of 8, and use the AdamW optimizer~\cite{loshchilov2018decoupled}. We use a linear learning rate warmup of 1,500 iterations and linear decay afterwards. During training, the source dataset is sampled using rare-class sampling~\cite{hoyer_daformer_2022} to address class imbalances.

When training UDA methods, we use a student-teacher setup, where the student's weights are aggregated during training into an EMA teacher model. The running weight for the EMA model is $\alpha=0.999$. This EMA teacher model is never backpropagated and is only used for pseudo-label generation. When creating the pseudo labels, the target images are not augmented, but we use horizontal flip aggregation to create the final pseudo label to reduce labeling noise. The threshold on the softmax outputs to generate the final pseudo labels is $\rho=0.968$ . We use a mask ratio of $r=0.7$, like in the original MIC~\cite{mic_2023}. However, diverging from MIC's strategy of masking image regions directly, our adaptation involves masking patch tokens when using a ViT encoder. The target images are mixed with the source images using DACS~\cite{Tranheden_2021_WACV} data augmentation. The final VFM-UDA method does not use the FD loss or HRDA, see \cref{sec:results:components}.

Our experimental setup aims to investigate the following performance aspects:
\begin{itemize}
    \item \textbf{VFM-UDA vs.~existing UDA methods:} We assess the performance of VFM-UDA against current UDA methods in synthetic, real, in-target, and out-of-target settings, focusing on segmentation quality. This wide range of test scenarios gives insights into each UDA method's generalization capabilities, crucial for real-world deployment.
    \item \textbf{Ablation on UDA components:} We conduct an in-depth evaluation of the individual impact and contributions of various UDA components within the VFM-UDA framework. This includes examining the effects of resolution adjustments, masking strategies, FD, and HRDA.
    \item \textbf{Efficiency analysis:} We compare the inference speed of VFM-UDA to that of existing UDA methods.
    \item \textbf{Exploring various VFMs:} We assess the effect of using different VFMs to find the relative advantages of various VFM pre-training strategies on the in- and out-of-target performance.
    \item \textbf{Impact of model size and pre-training:} We study how scaling up models with ImageNet and VFM pre-training affects the in- and out-of-target performance. This setup provides insights into how we can further scale UDA with larger model sizes.
\end{itemize}

\section{Results}
\label{sec:experiments}

\begin{table*}[ht!]
\centering
\begin{adjustbox}{width=\linewidth}

\begin{tabular}{lccccHcHccHcHc}
\toprule
 \multirow{2}[2]{*}{Method} & \multirow{2}[2]{*}{Backbone} &   \multirow{2}[2]{*}{Pre-training}& \multirow{2}[2]{*}{\# Parameters} & \multicolumn{5}{c}{In-target mIoU} &  \multicolumn{5}{c}{Out-of-target mIoU} 
 \\
 \cmidrule(lr){5-9}  \cmidrule(lr){10-14} 
 & &  & &Baseline& $\Delta$base& UDA& $\Delta$ oracle &  Oracle& Baseline &  $\Delta$base &  UDA &  $\Delta$ oracle &  Oracle \\
\midrule
DaFormer~\cite{hoyer_daformer_2022} & MiT-B5 & ImageNet-1K~\cite{Russakovsky_imagenet_2015}& 85.2M  & \light{47.1} & 21.2 & 68.3 & 8.3  & \light{76.6} & \light{41.9} & 8.2  & 50.1 &  0.3  & \light{66.8}\\
SePiCo~\cite{xie_sepico_2022} & MiT-B5 & ImageNet-1K~\cite{Russakovsky_imagenet_2015} &  85.2M      & \light{46.5} & 21.7 & 70.3 & 10.1 & \light{78.3} & \light{37.7} & 9.7  & 47.5 &  2.5  & \light{64.8}\\
HRDA~\cite{hoyer_hrda_2022} & MiT-B5 &  ImageNet-1K~\cite{Russakovsky_imagenet_2015} &  85.7M   & \light{47.3} & 25.9 & 73.8 & 7.5  & \light{80.8} & \light{40.6} & 10.2 & 50.8 &  1.1  & \light{66.9}\\
MIC~\cite{mic_2023} & MiT-B5 & ImageNet-1K~\cite{Russakovsky_imagenet_2015}  &     85.7M            & \light{47.3} & 28.1 & 75.9 & 5.4  & \light{80.8} & \light{40.6} & 14.6 & 55.2 &  -3.4 & \light{66.9}\\ 
VFM-UDA & ViT-B/14 & DINOv2~\cite{dinov2_2023}   &    88.4M                    & \light{62.9} & 13.1 & 77.1 & 6.5  & \light{82.4} & \light{60.4} & 2.6  & 61.3 &  11.8 & \light{74.8}\\
VFM-UDA & ViT-L/14 &  DINOv2~\cite{dinov2_2023}   &     307.6M                 & \light{68.7} &      & \textbf{79.0} &      & \light{83.4} & \light{64.8} &      & \textbf{65.5} &       & \light{76.3}\\

\bottomrule
\end{tabular}
\end{adjustbox}
\caption{\label{tab:syn2real} \textbf{Semantic segmentation performance for synthetic-to-real scenario.} We use the GTA5 $\rightarrow$ Cityscapes setup, a common benchmark for UDA methods. This scenario evaluates the performance of the UDA methods when there is a significant domain gap between the source and the target domain. The out-of-target dataset is WildDash2.}

\end{table*}

\begin{table*}[ht!]
\centering
\begin{adjustbox}{width=\linewidth}

\begin{tabular}{lccccHcHccHcHc}
  \toprule
 \multirow{2}[2]{*}{Method} & \multirow{2}[2]{*}{Backbone} &   \multirow{2}[2]{*}{Pre-training} &\multirow{2}[2]{*}{\# Parameters} &\multicolumn{5}{c}{In-target mIoU} &  \multicolumn{5}{c}{Out-of-target mIoU} 
 \\
 \cmidrule(lr){5-9}  \cmidrule(lr){10-14} 
 & &  & & Baseline& $\Delta$base& UDA& $\Delta$ oracle &  Oracle& Baseline &  $\Delta$base &  UDA &  $\Delta$ oracle &  Oracle \\
\midrule
    DaFormer~\cite{hoyer_daformer_2022} & MiT-B5  & ImageNet-1K~\cite{Russakovsky_imagenet_2015}& 85.2M  & \light{60.1} & 2.0& 62.1 & 8.5 & \light{70.7} & \light{51.7} & 0.5  & 52.2 & 14.6 & \light{66.8} \\
    SePiCo~\cite{xie_sepico_2022} & MiT-B5     & ImageNet-1K~\cite{Russakovsky_imagenet_2015} & 85.2M    & \light{60.8} & 1.6& 62.5 & 8.4 & \light{70.9} & \light{49.6} & 4.2  & 53.8 & 11.0 & \light{64.8} \\
    HRDA~\cite{hoyer_hrda_2022} & MiT-B5    & ImageNet-1K~\cite{Russakovsky_imagenet_2015} &  85.7M     & \light{64.4} & 5.5& 69.9 & 7.5 & \light{77.4} & \light{46.3} & 14.4 & 60.7 & 6.2  & \light{66.9}  \\
    MIC~\cite{mic_2023} & MiT-B5   & ImageNet-1K~\cite{Russakovsky_imagenet_2015} &   85.7M      & \light{64.4} & 8.9& 73.3 & 4.1 & \light{77.4} & \light{46.3} & 15.6 & 61.9 & 5.0  & \light{66.9}  \\
    VFM-UDA & ViT-B/14  & DINOv2~\cite{dinov2_2023}  &     88.4M                   & \light{75.7} & 2.3& \textbf{79.1} & 2.8 & \light{80.8} & \light{66.7} & 2.2  & 69.7 & 5.9  & \light{74.8}  \\
    VFM-UDA & ViT-L/14  & DINOv2~\cite{dinov2_2023}  &     307.6M            & \light{78.3} &    & \text{78.7} &     & \light{82.5} & \light{69.9} &      & \textbf{70.2} &      & \light{76.3}  \\

\bottomrule
\end{tabular}
\end{adjustbox}
\caption{\label{tab:real2real} \textbf{Semantic segmentation performance for the real-to-real scenario.} We use the Cityscapes $\rightarrow$ Mapillary setup, where both the source and the target domain contain real-world images, to analyze the performance of the UDA methods when the domain gap is relatively small. The out-of-target dataset is WildDash2.}

\end{table*}

In this section, we present the results and discuss the five experiments that we introduced in Sec.~\ref{sec:method}.

\subsection{Generalization of UDA with VFMs}
\label{sec:genstudy}

\paragraph{Overall findings.} The results of both the synthetic-to-real and real-to-real adaptation scenarios can be seen in Tab.~\ref{tab:syn2real} and Tab.~\ref{tab:real2real}, respectively. In this experiment, we only consider the VFM-UDA method with ViT-B/14, since it has a similar parameter count as MIC. On both UDA benchmarks, VFM-UDA demonstrates superior in-target and out-of-target performance compared to the current state-of-the-art UDA method, MIC. In the synthetic-to-real scenario, VFM-UDA adapts better than MIC by +1.2 mIoU points and generalizes better by +6.1 points. In the real-to-real one, the integration surpasses MIC even more, with differences of +5.8 mIoU in-target and +7.8 out-of-target. 

These results show that the generalization capabilities of VFMs and UDA methods are complementary, as the VFM-UDA combination achieves better UDA performance than the state-of-the-art UDA methods, while maintaining -- or even slightly improving -- the out-of-target performance of the VFM.

\paragraph{Effect of model size.} When integrating UDA with a significantly larger model, ViT-L/14, the adaptation and generalization capabilities of the model increase even more, outperforming the state-of-the-art UDA method by larger margins. In the real-to-real scenario, it is interesting to note that the combination VFM-UDA yields only a minor performance increase compared to its VFM baseline, both in-target and out-of-target. Essentially, when the VFM is large, the added benefits from UDA to the model's overall performance become marginal. This suggests that the performance improvements obtained by scaling VFMs might limit the additional generalization benefits achievable by pairing with UDA techniques, especially in simpler settings like real-to-real scenarios. For a more in-depth analysis of the scalability of VFM-UDA with different pre-trainings, we refer to Sec.~\ref{sec:scalability}. 

Next, we will also demonstrate that these larger models can be faster than the smaller state-of-the-art UDA method.

\subsection{Evaluation of UDA components}
\label{sec:results:components}
To investigate how each UDA component affects the overall adaptation performance when integrated with VFMs, we use the synthetic-to-real adaptation scenario. The baseline UDA method only applies supervised learning to the source domain and self-training to the target domain. Using this baseline, we try to incrementally improve it by introducing the following configurations: 

\begin{itemize}
    \item \textbf{Incorporation of masking:} adding Mask Image Consistency (MIC) when performing self-training, either at image or token level.
    \item \textbf{Feature Distance (FD):} adding the FD loss to prevent the model from forgetting the pre-training.
    \item \textbf{Multi-resolution training:} applying multi-resolution training by fusing high-resolution and low-resolution predictions, as proposed in HRDA~\cite{hoyer_hrda_2022}.
\end{itemize}

\begin{table}
\centering
\begin{adjustbox}{width=0.85\linewidth}
    \begin{tabular}{cccccc}
    \toprule
        \thead{Pseudo \\ labeling} & \thead{Image \\ masking} & \thead{Token \\ masking} & FD & HRDA & mIoU  \\ 
        \midrule
          -- & -- & -- & -- & -- & 62.9    \\ 
          $\checkmark$ & -- & -- & -- & -- & 72.8    \\ 
          $\checkmark$ & $\checkmark$ & --  & -- & -- & 76.8  \\  
           \rowcolor{gray!30}
          $\checkmark$ & -- & $\checkmark$  & -- & -- & \textbf{77.1}  \\  
          $\checkmark$ & -- & $\checkmark$ & $\checkmark$ &  -- & 75.1  \\ 
          $\checkmark$ & -- & $\checkmark$  & -- & $\checkmark$ & 75.1  \\ 
          $\checkmark$ & -- & $\checkmark$  & $\checkmark$ & $\checkmark$ & 76.9  \\ 
     \bottomrule
    \end{tabular}
\end{adjustbox}
\caption{\label{tab:udacomp} \textbf{Analysis of VFM with different UDA components.} The model is initialized with DINOv2 pretraining, adapting GTA5 $\rightarrow$ Cityscapes. Using masking improves performance, while the multi-resolution training proposed in HRDA~\cite{hoyer_hrda_2022} has no positive effect.}
\end{table}

\begin{table}[ht!]
\centering
\begin{adjustbox}{width=\linewidth}
    \begin{tabular}{ccccc}
    \toprule
        Method & \thead{Image \\ size (px)} & mIoU & \thead{Time \\ (ms)} & Speedup  \\ 
        \midrule
        
       \light{MIC (MiT-B5)}  & \light{1024$\times$2048} & \light{75.9} & \light{1072} & \light{1.0$\times$} \\   
        \midrule
        DaFormer (MiT-B5) & 512$\times$1024 & 68.3 & 110 & 9.7$\times$   \\ 
        SePiCo (MiT-B5)& 640$\times$1280 & 70.3  & 116  & 9.3$\times$  \\ 
        HRDA (MiT-B5)  & 1024$\times$2048 & 73.8 & 1072 & 1.0$\times$ \\         
        MIC (MiT-B5) & 1024$\times$2048 & 75.9 & 1072 & 1.0$\times$ \\   
       
        VFM-UDA (ViT-B/14)& 1024$\times$2048 & 77.1 & 128 & 8.4$\times$ \\ 
        VFM-UDA (ViT-L/14) & 1024$\times$2048 & 79.0 &  323 & 3.3$\times$ \\ 
    \bottomrule
    \end{tabular}
\end{adjustbox}
\caption{\label{tab:speed}\textbf{Inference runtime analysis.} After adapting GTA5 $\rightarrow$ Cityscapes, the performance of all UDA methods is measured on an Nvidia A6000 GPU with 16-bit mixed precision. The VFM-UDA combination benefits from higher inference speed compared to its baseline MIC, even when using ViT-L/14 which has 3.6$\times$ more parameters.}

\end{table}

\paragraph{Findings.} Our analysis, detailed in Tab.~\ref{tab:udacomp}, reveals nuanced performance impacts for each UDA component. Token Masking, as opposed to the original Image Masking used in MIC, yields a slightly better performance. When we use either the FD loss or HRDA on top of Token Masking, there is a noticeable decline in mIoU. This suggests that these components may not translate as effectively to ViT-based encoders, which lack hierarchical features that are present in the MiT-B5-encoder-based UDA methods. Although the full combination of Token Masking, the FD loss, and HRDA shows some improvement over using them separately, the combined effect still does not exceed the performance of the Token Masking alone. These findings imply that the FD and the HRDA components may be redundant when using ViT-based VFMs. Therefore, we do not incorporate them in the final VFM-UDA method.

\paragraph{Inference speed findings.} Tab.~\ref{tab:speed} shows the inference speed of the methods compared in Tab.~\ref{tab:syn2real}. The VFM-UDA approach shows a large improvement in inference speed  Specifically, using the ViT-B/14 model achieves an 8.4$\times$ speed increase over the prior state-of-the-art method MIC~\cite{mic_2023}. It obtains this speed because -- unlike the HRDA and MIC methods -- it does not use the HRDA technique that requires multiple inference passes.
Even when scaling up to a ViT-L/14 model, which has 3.6$\times$ more parameters, the VFM-UDA method still maintains a significant advantage, operating 3.3$\times$ faster than HRDA-based approaches. The experiments are run on a Nvidia A6000 GPU with a batch size of 1, and we report the average inference time per image.

\subsection{Analysis with different VFMs}
\label{sec:results:vfms}
To determine the most effective Vision Foundation Model (VFM) for UDA, we initially selected DINOv2 due to its robust performance across a range of downstream tasks~\cite{dinov2_2023}. To validate this choice and explore alternatives, we extend our analysis to include two other VFMs, EVA02~\cite{eva02_2023} and EVA02-CLIP~\cite{Sun_eva02clip_23}. These models were chosen for their close performance to DINOv2, making them suitable candidates for comparison~\cite{dinov2_2023}. We evaluated these VFMs in a synthetic-to-real adaptation scenario, specifically adapting GTA5 to Cityscapes, and assessed both in-target and out-of-target performance to study their adaptation and generalization capabilities.

\paragraph{Findings.} 
Tab.~\ref{tab:vfm} shows that DINOv2 consistently outperforms EVA-02 and EVA-02-CLIP with a significant margin of +4.8 mIoU points in terms of in-target performance and +2.9 points in terms of out-of-target performance compared to EVA-02-CLIP. While EVA-02 and EVA-02-CLIP yield similar results in out-of-target scenarios, EVA-02-CLIP surpasses EVA-02 in terms of in-target performance. This experiment underscores DINOv2's superior adaptability and generalization, supporting its selection for the VFM-UDA method. 

\begin{table}[t]
\centering
\begin{adjustbox}{width=\linewidth}
    \begin{tabular}{ccc}
    \toprule
        Method & In-target (mIoU) & Out-of-target (mIoU)  \\ 
        \midrule
        EVA-02~\cite{eva02_2023} & 65.8 & 57.3 \\ 
        EVA-02-CLIP~\cite{Sun_eva02clip_23} & 72.3 &  58.4 \\
        DINOv2~\cite{dinov2_2023} & 77.1 & 61.3  \\ 
    \bottomrule
    \end{tabular}
\end{adjustbox}
\caption{\label{tab:vfm} \textbf{Effect of using different VFMs}. When adapting GTA5 $\rightarrow$ Cityscapes, on the ViT-B/14 encoder, DINOv2~\cite{dinov2_2023} provides the best in-target and out-of-target performance.}
\end{table}

\subsection{Effect of model size with different pre-training strategies}
\label{sec:scalability}
In our examination of model scaling with ImageNet pre-training versus VFM pre-training, we compare the adaptation and generalization capabilities at varying sizes: small (ViT-S/14), base (ViT-B/14), and large (ViT-L/14). We use DeiT III~\cite{deit3_22} for the ViTs pre-trained on ImageNet-1K.

\begin{table}[t]
\centering
\begin{adjustbox}{width=\linewidth}
    \begin{tabular}{ccccc}
    \toprule
         \multirow{2}[2]{*}{Method} & \multicolumn{2}{c}{In-target (mIoU)} & \multicolumn{2}{c}{Out-of-target (mIoU)}  \\ 
         \cmidrule(lr){2-3} \cmidrule(lr){4-5} 
         & ImageNet-1K & DINOv2 & ImageNet-1K & DINOv2  \\
        \midrule
        ViT-S/14  & 66.9 & 69.7 & 46.0 & 56.2 \\ 
        ViT-B/14  & 68.1 & 77.1 & 51.8 & 61.3 \\ 
        ViT-L/14  & 67.3 & 79.0 & 54.6 & 65.5 \\ 
    \bottomrule
    \end{tabular}
\end{adjustbox}
\caption{\label{tab:pretraining} \textbf{Effect of model size with different pre-training strategies.} We use an ImageNet pre-trained ViT as our non-VFM model and DINOv2 as the VFM. Adapting GTA5 $\rightarrow$ Cityscapes and scaling with model size, the ImageNet pre-trained model is unable to scale in performance, while the DINOv2 shows a consistent increase for both in- and out-of-target.}
\end{table}

\paragraph{Findings.} Tab.~\ref{tab:pretraining} shows that models pre-trained on ImageNet do not exhibit improved performance with increased model size in the in-target setting. In contrast, models pre-trained with DINOv2 show a consistent improvement in performance as model size increases, for both in-target and out-of-target evaluations. This shows the benefit of using VFMs for UDA, as they have the potential to benefit from increased model scale, to achieve superior generalization.

\section{Conclusions}
\label{sec:conclusions}
In this work, we explore whether the generalization capabilities of UDA and VFMs are complementary, to obtain models that can excel at both adaptation to a specific target domain and generalization beyond this target domain. Due to architectural differences between VFMs and encoders previously used in UDA, we made the necessary adjustments for the combined model to work in multiple configurations. From the experiments, we found that at equivalent model sizes, the combined VFM-UDA model can (a) adapt better to target domains than current state-of-the-art UDA methods, while (b) maintaining -- or even slightly improving -- the out-of-distribution generalization performance of VFMs. Moreover, we found that the VFM-UDA combination benefits from increased model scale, as larger VFMs yield higher in- and out-of-target performance. This study sets new standards and baselines for UDA for target-specific adaptation and out-of-distribution generalization, and offers practical guidelines for integrating VFMs into UDA to harness their joint benefits.

\paragraph{Acknowledgment} This work has received funding from Chips Joint Undertaking (Chips JU), under grant agreement No 101097300. The Chips JU receives support from the European Union’s Horizon Europe research and innovation program and Austria, Belgium, France, Greece, Italy, Latvia, Netherlands, Norway. 

The project Ecomobility has received funding within the Chips Joint Undertaking (Chips JU), a public-private partnership in collaboration with the Horizon Europe (HORIZON) Framework Programme and National Authorities under grant agreement number 101112306.

This work made use of the Dutch national e-infrastructure with the support of the SURF Cooperative using grant no. EINF-7020, which is financed by the Dutch Research Council (NWO).

{
    \small
    \bibliographystyle{ieeenat_fullname}
    \bibliography{main}
}

\end{document}